\documentclass[10pt,times,mathptm,psfig,twocolumn,journals]{IEEEtran}
\usepackage{url}
\usepackage{amssymb}
\usepackage{array}
\usepackage{amsmath}
\usepackage{graphicx}
\usepackage{booktabs}
\usepackage{epsfig}
\usepackage{ulem}
\usepackage{hyperref}
\hypersetup{pdfborder={0 0 0}, colorlinks=true, linkcolor=red, citecolor=blue}
\usepackage{subfigure}
\usepackage{cite}
\usepackage{epstopdf}
\usepackage{amsmath}
\usepackage[misc]{ifsym}
\usepackage{bm}
\usepackage{fancyhdr}
\usepackage{mathrsfs}
\usepackage{bbding}
\usepackage{algorithm}
\usepackage{algorithmic}
\usepackage{setspace}
\usepackage{multirow}
\usepackage{tabularx}
\usepackage{url}
\usepackage{lipsum}
\usepackage{booktabs}
\usepackage[bottom]{footmisc}

\begin{document}

\title{Neural Contourlet Network for Monocular 360° Depth Estimation}


\author{Zhijie~Shen, Chunyu~Lin,~\IEEEmembership{Member,~IEEE}, 

Lang Nie,
Kang~Liao, Yao~Zhao,~\IEEEmembership{Senior Member,~IEEE}






\thanks{This work was supported by the National Natural Science Foundation of China (No. 62172032, U1936212, 62120106009).
\textit{(Corresponding author: Chunyu Lin)}}
\thanks{Zhijie Shen, Chunyu Lin, Lang Nie, Kang Liao, Yao Zhao are with the Institute of Information Science, Beijing Jiaotong University, Beijing 100044, China, and also with the Beijing Key Laboratory of Advanced Information Science and Network Technology, Beijing 100044, China (email: zhjshen@bjtu.edu.cn, cylin@bjtu.edu.cn, nielang@bjtu.edu.cn, kang\_liao@bjtu.edu.cn, yzhao@bjtu.edu.cn)).}}

\maketitle

\begin{abstract}
     For a monocular 360° image, depth estimation is a challenging because the distortion increases along the latitude. To perceive the distortion, existing methods devote to designing a deep and complex network architecture. In this paper, we provide a new perspective that constructs an interpretable and sparse representation for a 360° image. Considering the importance of the geometric structure in depth estimation, we utilize the contourlet transform to capture an explicit geometric cue in the spectral domain and integrate it with an implicit cue in the spatial domain. 
     Specifically, we propose a neural contourlet network consisting of a convolutional neural network and a contourlet transform branch. In the encoder stage, we design a spatial–spectral fusion module to effectively fuse two types of cues. Contrary to the encoder, we employ the inverse contourlet transform with learned low-pass subbands and band-pass directional subbands to compose the depth in the decoder.
     Experiments on the three popular 360° panoramic image datasets demonstrate that the proposed approach outperforms the state-of-the-art schemes with faster convergence. Code is available at \url{https://github.com/zhijieshen-bjtu/Neural-Contourlet-Network-for-MODE}.
\end{abstract}
\markboth{IEEE Transactions on Circuits and Systems for Video Technology}
\IEEEpeerreviewmaketitle

\section{Introduction}
\label{section1}
\IEEEPARstart{R}{esearchers} favor monocular omnidirectional depth estimation tasks because of its larger FoV(Field of View) and low cost. However, it is undeniable that inferring relevant 3-D information from 2-D images is an ill-posed problem. Moreover, distortions in panoramic images seriously hinder the performance of depth estimation models. Most of the existing solutions focus on attenuating the negative effects of distortions by employing various distortion-aware convolution filters or adopting a projection-fusion strategy. However, these distortion-specific designs make the networks much more complicated and burdened. Furthermore, the performance heavily relies on the implicit depth-related cue extracted from CNNs, which lacks of interpretability and slows down the convergence of the network.

In contrast to learning-based schemes, traditional depth estimation methods mainly focus on discovering different cues to estimate depth from 2-D images~\cite{8353313,9615228,8764412,8105853}, such as structure, shadows, lighting, occlusion, etc~\cite{dijk2019neural}. However, these cues are limited or only applicable to specific scenarios (e.g., shadows are greatly affected by ambient lighting). It is interpretable but not robust for various scenes. 
In this paper, we rethink the panoramic depth estimation task from the perspective of beneficial cues, putting forward a simple but effective solution combining the learning-based scheme with traditional explicit depth-related cues. 
  
\paragraph{Which cue is needed?}
To keep our framework as straightforward as possible, we cannot take all of the above depth cues into account, so we must find the most suitable ones. Luckily, previous works have confirmed that geometric structure is one of the most essential cues used by depth estimation. Specifically, Chen et al.\cite{Chen_2021_CVPR} gain better performance by learning an effective structure representation. Shen et al.\cite{shen2021distortion,Pintore_2021_CVPR} improve the accuracy discernibly by constraining the depth edges. Apparently, the structure cue is reasonable. Considering the human visual system (HVS), biologists generally believe that central vision is highly susceptible to the influence of the limbic vision~\cite{sivak1990integration}. That is why humans always get annoyed by the ads popping from the corner while enjoying the movies. Geometric structures are a significant part of limbic vision and cannot be ignored in most pixel-level vision tasks \cite{8307430}.
  
\paragraph{How to catch omnidirectional geometric cue?}
CNNs can extract abundant semantic features but also lacks the ability of geometric transformation~\cite{MengkunLiu2021CCNNCC}, especially under the challenge of large distortion in panoramas. It means that the implicit cue extracted by CNN may contain geometric cue, but it is too weak to help CNN predict omnidirectional depths.
Inspired by~\cite{ramamonjisoa2021single}, we introduce wavelet analysis tools to assist the CNN branch in catching the geometric cue. Wavelet transformation can decompose an image into low-pass subbands and high-frequency coefficient subbands (implicit geometric structures). Furthermore, the wavelet transform offers a multi-resolution structure that can provide panoramic structural features of different resolutions. However, unlike perspective images, the distortion in panoramas yields curvilinear structures more efficiently. Directly applying the general wavelets (e.g., Haar Wavelet) is challenging to capture the curves perfectly.
To this end, we propose to adopt contourlet to perceive the omnidirectional geometric cue for its beneficial properties of multi-resolution, multi-direction, and efficiency in curve approximation.
  
In this paper, we propose a simple but effective neural contourlet network for panoramic monocular depth estimation. 
The network is composed of a CNN and a contourlet branch to capture an implicit cue and an explicit geometric cue, respectively. Specifically, the forward contourlet transform is employed in the encoder stage to decompose the input image into a series of band-pass directional subbands with geometric cues in the spectral domain. Furthermore, we propose the \textit{Spatial–Spectral Fusion Module} (SSFM) to integrate the explicit geometric cue with the implicit cue by bridging the domain gap between the spectral and spatial domains. In contrast to the encoder, inverse contourlet transform is utilized to compose the omnidirectional depth with a series of learned subbands in the decoder stage. In particular, we propose \textit{Coefficients Generation Module} (CGM) to guide the generation of subands 
to compose depth maps, building a learnable contourlet representation for the omnidirectional depth maps. Compared with other monocular omnidirectional depth estimation solutions, our framework is more interpretable, effective, and straightforward.
 
  
We validate the effectiveness of the proposed framework on three popular panoramic datasets (Stanford2D3D~\cite{DBLP:journals/corr/ArmeniSZS17}, Matterport3D~\cite{Chang2018Matterport3D}, and Structured3D~\cite{zheng2020structured3d}). The experimental results show that our method significantly outperforms other state-of-the-art solutions with a faster convergence rate. The contributions of this paper can be summarized as follows.
  
  \begin{itemize}
    
    
   \item We propose the neural contourlet network that combines an implicit cue with an explicit geometric cue for monocular omnidirectional depth estimation.
   
   \item We pioneer in employing contourlet transform to effectively capture the panoramic geometric cue for its ``panorama-customized" properties (e.g., sparse representation of curves, muti-resolution, multi-directionality), making our framework more interpretable and yielding a faster convergence. 
    
    \item We propose \textit{Spatial–Spectral Fusion Module} and \textit{Coefficients Generation Module} to bridge the domain gap between the contourlet and CNN branch and to provide learned subbands for composing omnidirectional depth maps, respectively.
    

    
\end{itemize}
The remainder of this paper is organized as follows. In
Sec. \ref{relatedwork}, we describe the related work of monocular 360° depth estimation and the application of wavelet in image processing. In Sec. \ref{method}, we introduce the overall model structure and describe each module in details. Finally, the experimental results are presented in Sec. \ref{experiment}.

\section{Related Work}
\label{relatedwork}
In this section, we will discuss the related works on omnidirectional depth estimation and wavelet in image processing.
\subsection{Monocular 360° Depth Estimation}
The focus of the monocular panoramic depth estimation task is to deal with panoramic distortion, and the current solutions can be roughly classified into three technical routes.
\subsubsection{Perspective Projection}
To address omnidirectional distortion in the equirectangular map, Wang et al. \cite{2020BiFuse} propose Bifuse to fuse cube and equirectangular features. Cube features with less distortion can assist in perceiving omnidirectional distortion. From a lightweight perspective, Jiang et al.~\cite{Jiang_2021} improve Bifuse by designing an efficient fusion block and propose Unifuse. Besides, Shen et al.~\cite{shen2021distortion} utilize a dual-cube fusion strategy to get rid of distortion. Compared with cubemap, there are less distortion and pixel loss in the tangent images, and Eder et al.~\cite{Eder2020Tangent} propose leveraging tangent images to mitigate the spherical distortion. These methods require converting projection formats frequently, resulting in instability in the training stage~\cite{Chen_2021}.
\subsubsection{Spherical Representation}
To deal with the distortion in panoramas, many researchers~\cite{tateno2018distortion,eder2019mapped} have focused on adjusting the sampling positions of standard convolution kernels and designed many distortion-aware filters, such as spherical-conv~\cite{cohen2018spherical} and equi-conv~\cite{FernandezLabrador2020CornersFL}. Essentially, both are dynamic convolutional structures with spherical coordinate prior guidance. They can effectively perceive panoramic distortion but inevitably increase the network’s computational burden, making them difficult to follow.
\subsubsection{Indoor Layout Prior}
Assuming the indoor sceneries are gravity aligned, Pintore et al.~\cite{Pintore_2021_CVPR} proposed SliceNet. However, it is unsuitable for non-gravity aligned areas (floor, ceiling). Joint with room layout, Jin et al.~\cite{Jin_2020_CVPR} presents a deep learning framework that leverages structure (points, liens, planes) for indoor depth estimation. But it requires additional supervision. Moreover, Zeng et al.~\cite{zeng2020joint} employ the layout depth map as an intermediate representation, achieving a better performance. Recently, Zioulis et al.~\cite{zioulis2022monocular} propose leveraging weak layout cues to estimate 360 depth.
\subsection{Wavelet in Image Processing}
Wavelet analysis is an effective tool for image processing. Recently, it has been combined with neural networks to solve computer vision tasks\cite{8116695}. In image classification tasks, ~\cite{QiufuLi2020WaveletIC} improve noise-robustness by replacing pooling operation with discrete wavelet transform. Super-resolution~\cite{guo2017deep,deng2019wavelet} solutions apply inverse wavelet transform to generate high-resolution RGB images. Similarly, Liu et al.~\cite{MengkunLiu2021CCNNCC} introduce contourlet transform to image classification task by extracting features in the spectral domain. However, extracting features entirely in the spectral domain will weaken depth-related cues. 
Ramamonjisoa et al.~\cite{ramamonjisoa2021single} predict the depth map from a single perspective image by designing more sparse wavelet coefficient maps. However, wavelet transform is not suitable for panoramas that have much more curved structures. In contrast with all these works, we focus on utilizing contourlet transform for capturing the geometric cue for depth, dealing with the panoramic distortions, and presenting a simple but effective framework for estimating 360° depth.

\begin{figure}[t]
\includegraphics[width=0.45\textwidth]{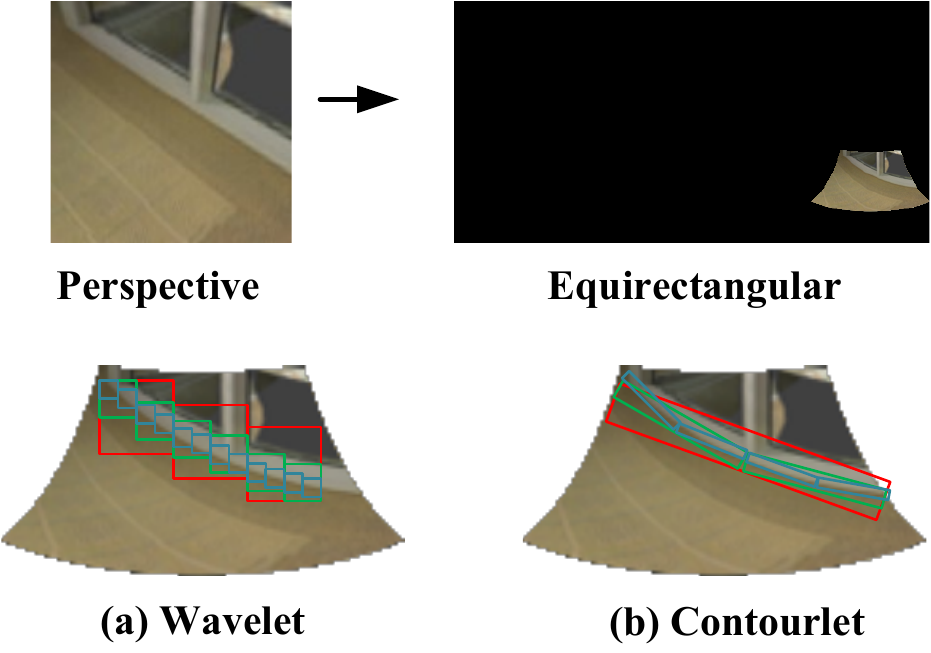}
\caption{Linear structures in perspectives are distorted in panoramas. Contourlet has elongated supports and can efficiently represent smooth contours with fewer coefficients. The square support of wavelets produces a  pronounced mosaic effect at depth edges (especially curved edges), as shown in Fig. \ref{fig:abla} d.}
\label{fig:sparse}
\end{figure}
\begin{figure*}[t]
\centering
\includegraphics[width=\textwidth]{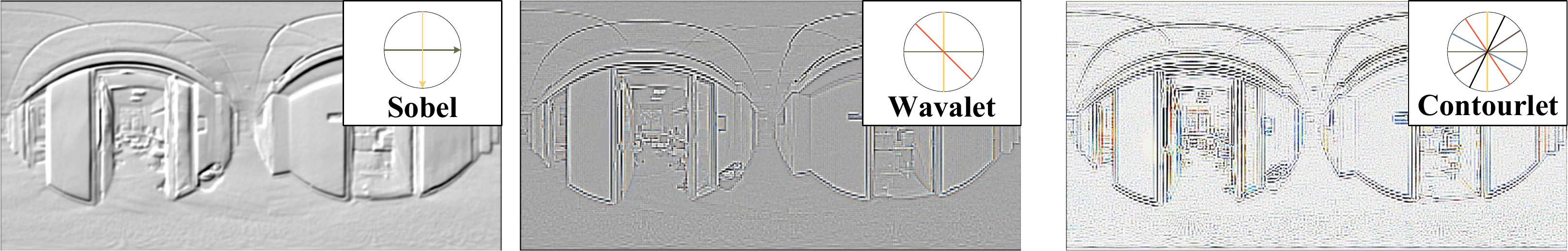}
\caption{Illustration of extracted panoramic structures. Sobel (left) operator has only two extraction directions, wavelet (middle) has three and contourlet (right) has eight.}
\label{fig:directional}
\end{figure*}
\section{Method}
\label{method}
In this section, we discuss the properties of contourlet (the motivation, in Sec. \ref{motivation}) firstly; then introduce the overall model structure (in Sec. \ref{NCN}) and describe each module in detail (in Sec. \ref{CNNbranch}, \ref{contourletbranch}, \ref{SSFM}, \ref{CGM} and \ref{AF}); finally, we describe our objective function (in Sec. \ref{objective}). 

\subsection{Why contourlet?}
\label{motivation}
Existing deep learning-based solutions for panoramic depth estimation focus on removing the negative effects of distortions. However, dealing with distortions is a challenging task that would seriously increase the complexity of the network. Motivated by traditional depth estimation solutions that leverage depth-related cues to infer the depth, we propose to deal with the distortion by exploring a helpful geometric cue.

Considering the abundant curved structures in panoramas, we utilize contourlet to extract the panoramic geometric cue for its multi-resolution, sparsity of representing curves, and multi-directionality. Next, we discuss and prove the ``panorama-customized" properties of the contourlet below.

Firstly, as an extension of the wavelet analysis tool, contourlet has the advantage of multi-resolution property and can provide optimal geometric cues in spectral-domain for different resolutions of the CNN branch. Secondly, the contourlet has elongated support space (Fig. \ref{fig:sparse}(b)), which can linearly approximate the curved structure with sparse coefficient representation. However, wavelet has square support (Fig. \ref{fig:sparse}(a)) space that approximates the curves by point with much more coefficients, resulting in a mosaic effect (see Fig. \ref{fig:abla} d) at depth edges. 
Finally, a prominent feature of contourlet is multi-directionality (Fig. \ref{fig:directional}(right)), which can assist in perceiving omnidirectional geometric cues effectively. 
Then, we give brief proofs for its sparsity and multi-directionality.

\textbf{Sparsity of representing curves.} Given a 2-D smooth curve equation $f$ (continuously differentiable of 2 order), a series expansion of $f$ can be expressed as
\begin{equation}
f = \sum_{i = 1}^{\infty}\alpha_{i}\phi_{i}   
\end{equation}
where $\alpha_{i}$ is denoted as coefficients and $\left \{ \phi  \right \} _{i=1}^{\infty }$ is the basis. We use the squared ($L_{2}$-norm square) error of the M-term expansion to measure the sparsity, which can be written as:
\begin{eqnarray}
\left \| f-\hat{f}_{M}^{\left \{ \phi  \right \} _{i=1}^{\infty }} \right \|\le  C(\log_{}{M} )^{3}M^{-v}
\end{eqnarray}
where $\hat{f}_{M}^{\left \{ \phi  \right \} _{i=1}^{\infty }}$ denotes the best $M$-term approximation of function $f$ with the basis $\left \{ \phi  \right \} _{i=1}^{\infty }$; $M$ is the number of the most significant coefficients in $\left \{\alpha_{i}\right \}$; $v$ and $C$ are constants. From \cite{MengkunLiu2021CCNNCC}, we can get
\begin{eqnarray}
\left \| f-\hat{f}_{M}^{(Wavelet)} \right \|\asymp M^{-1}.
\end{eqnarray}
\begin{eqnarray}
\left \| f-\hat{f}_{M}^{(Contourlet)} \right \|\asymp M^{-2}.
\end{eqnarray}
According to \cite{DavidLDonoho1998DataCA}, we can tell that contourlet has the optimal approximation rate ($M^{-2}$) of curve signal $f$ than wavelet ($M^{-1}$), which means contourlet achieves the best sparse representation of curves.

\textbf{Mutidirectionality.}
In fact, many traditional methods can be used to extract geometric structures, but not all of them are suitable for omnidirectional images (panoramas). As shown in Fig. \ref{fig:directional}, the sobel operator has only two directions (horizontal and vertical) that are not enough for the omnidirectional images. In addition, the Sobel operator does not have the advantage of multi-resolution analysis property that cannot naturally fit the hierarchical architecture of the CNN branch. Wavelet is a commonly used multi-resolution analysis tool, but it has only three directions (horizontal, vertical, and diagonal) and cannot optimally represent omnidirectional structures. In contrast, the contourlet transform is hierarchical and can decompose an image into $2^{l}$ directional subbands by directional filter bank \cite{MengkunLiu2021CCNNCC}.
\begin{figure}[t]
\centering
\includegraphics[width=0.45\textwidth]{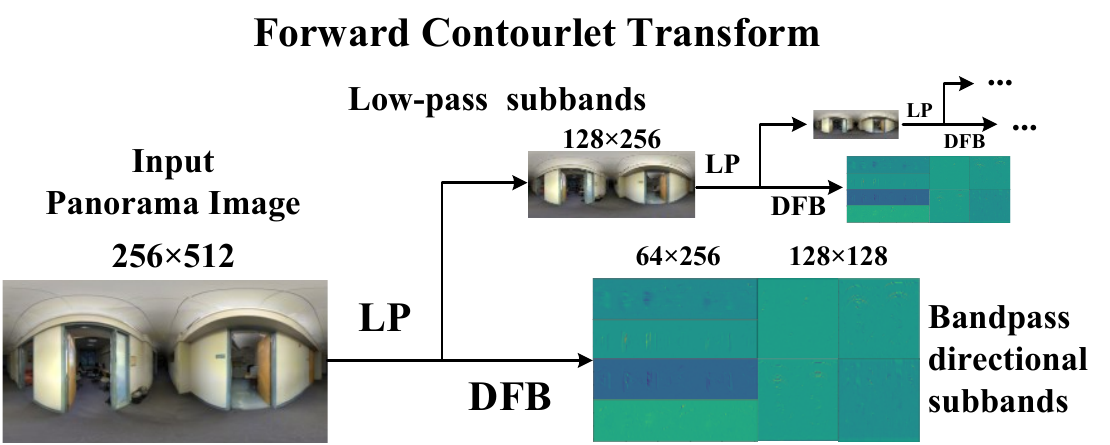}
\caption{Each contourlet transformation can decompose a 256×512 panorama image to a 128×256 low-pass subband and group of bandpass directional subbands (eight, four with size 64×256, and four with 128×128).}
\label{fig:FCT}
\end{figure}
\begin{figure*}[t]
\centering
\includegraphics[width=\textwidth]{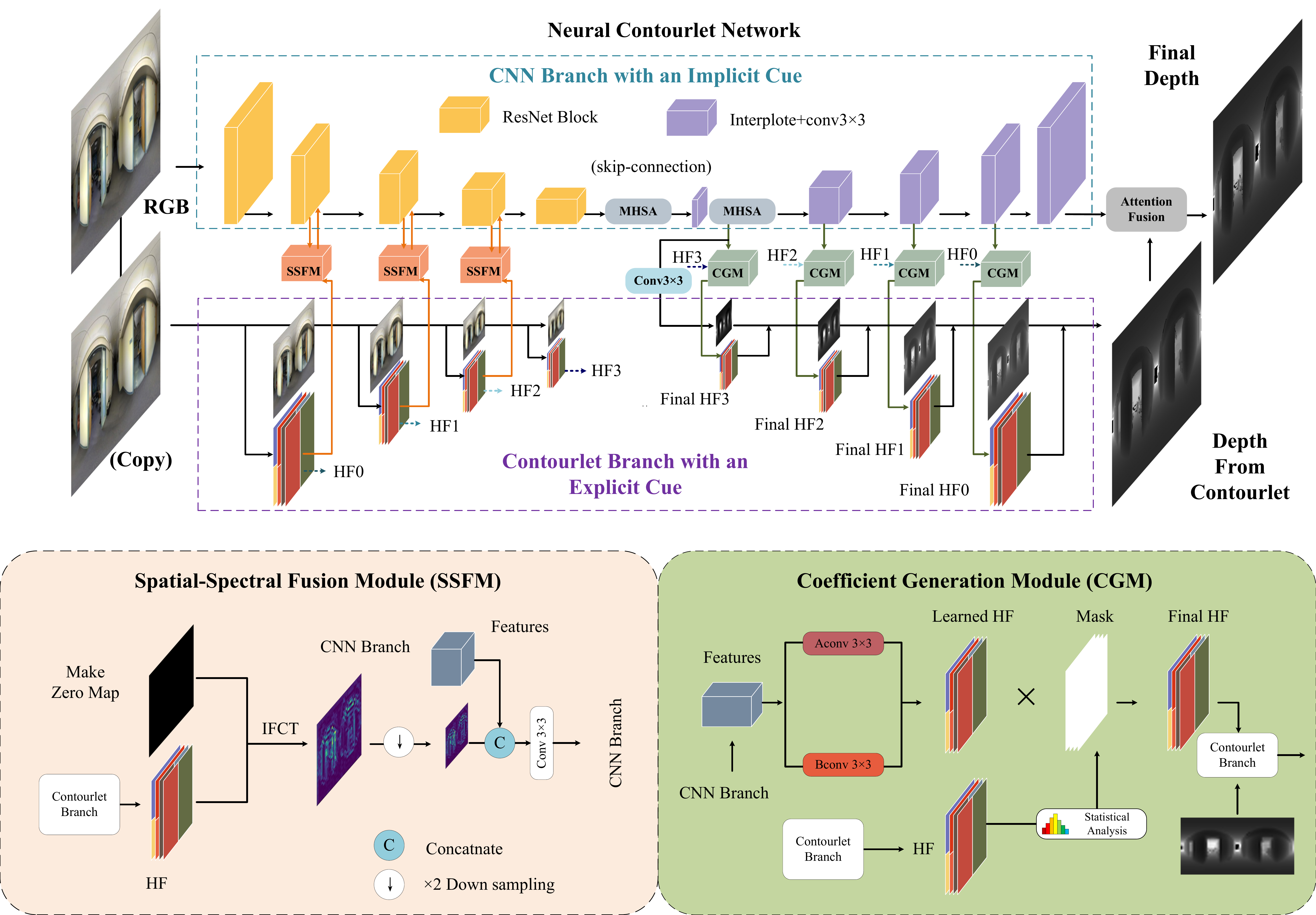}
\caption{Neural contourlet network. $HF$ or $HF_{s}, s\in {1, 2, 3, 4}$, represent the coefficients used in contourlet transform. $IFCT$ denotes the inverse contourlet transform that is intended to generate size-matched coefficient maps. In the \textit{Spatial–Spectral Fusion Module}, due to the mismatch in the size of the reconstructed panoramic structure map, a 2-fold downsampling is required and then concatenated with the feature map of the CNN branch. In \textit{Coefficients Generation Module}, $Aconv$ is implemented by setting the stride of standard conv 3×3 to (2, 1), and the stride in $Bconv$ is (1, 2). Statistical analysis represent the our proposed mask generation algorithm and the visushrink threshold function mentioned in Sec. \ref{CGM}.}
\label{fig:net}
\end{figure*}
\subsection{Neural Contoulet Network}
\label{NCN}
As shown in Fig. \ref{fig:net}, we propose a neural contourlet network by combining contourlet and convolutional neural networks. The network is a hierarchical architecture with two branches - the CNN branch and the contourlet branch. The CNN branch is employed to extract semantic features, and the contourlet branch is utilized to capture panoramic geometric cue. Concretely, the CNN branch consists of 3 parts: ResNet blocks, non-local (muti-head self-attention, MHSA) layers, and upsampling blocks. The contourlet branch mainly comprises multi-scale forward contourlet transform and its inverse contourlet transform. The proposed \textit{Spatial–Spectral Fusion Module} connects the two branches in the encoder stage, and the designed \textit{Coefficients Generation Module} to provide learned coefficients for composing depth maps in the decoder stage. 
In our network, the geometric cues from the contourlet branch is leveraged to assist the CNN branch in extracting discriminative semantic features. Subsequently, the extracted semantic features can be guided to generate coefficients with the geometric cue for composing depth maps.
The combination of implicit and explicit panoramic geometric cues improves the performance of the panoramic depth estimation model, yielding a faster convergence. The details of the network will be discussed in the Sec. \ref{CNNbranch}, \ref{contourletbranch}, \ref{SSFM}, \ref{CGM} and \ref{AF}.



\subsubsection{Convolutional Neural Network Branch }
\label{CNNbranch}
Given a single omnidirectional image input $E$ (represented in equirectangular projection and $E \in \mathbb{R}^{3\times H \times W}$), it is firstly 
fed to ResNet \cite{he2016identity} blocks (weights are pre-trained on ImageNet \cite{he2019rethinking}) to extract semantic features with resolution of 1/2, 1/4, 1/8, and 1/16, respectively. After that, the lowest resolution (1/16) features from ResNet blocks will pass through an MHSA layer, an up-sampling block, and another MHSA layer \cite{wang2018non} successively to establish the global perception field. Then, four up-sampling blocks are adopted to predict the final depth map in the decoder, where each up-sampling block \cite{ramamonjisoa2021single, huang2018condensenet} consists of the nearest interpolating layer and two 3×3 convolutional layers.
Moreover, skip-connection is employed to connect the low-level and high-level features with the same resolution in the CNN branch to prevent the gradient vanishing problem and information imbalance in training~\cite{ronneberger2015u}. 
\subsubsection{Contourlet Branch}
\label{contourletbranch}
Forward contourlet transform utilizes Laplacian pyramid to decompose a 2-D image into two parts: low-pass subbands and a high-pass subbands. Furthermore, the high-pass subbands will be sent to a directional filter bank for multidirectional analysis.  The directional filter bank will divide the high-frequency parts into $2^{l}$ directional subbands, of which subband $0$ to $2^{l-1}-1$ and $2^{l-1}$ to $2^{l}-1$ correspond to the vertical and horizontal details, respectively. Similar to~\cite{MengkunLiu2021CCNNCC}, we denote $l = 3$ in this paper, which means we get four vertical and four horizontal bandpass subbands (illustrated in Fig. \ref{fig:FCT}). In the decoder stage, inverse contourlet transform is employed to compose a depth map with learned subbands. However, these subbands captured in the contourlet branch do not match the dimensions of the features in the CNN branch. Besides, subbands and features come from different domains, and it is not reasonable to fuse them directly (e.g., add or concatenate). To address this issue, we propose the \textit{spatial–spectral fusion module} to bridge the domain gap and match the dimensions.
\subsubsection{Spatial–Spectral Fusion Module}
\label{SSFM}
To bridge the gap between the spatial and spectral domain, we transform the bandpass directional subbands to the spatial domain by inverse contourlet transform and then integrate the bi-branch cues. 
Specifically, for each resolution $s$, the features $f_{fusion}^{s}$ that are fed to CNN branch can be expressed as:
\begin{equation} \label{eqn2}
  \begin{split}
  f_{fusion}^{s} &= Conv(Concat(\mathcal{T}^{-1} (f_{h}^{fct,s}), f^{cnn,s})),  
  \end{split}
\end{equation}

where $\mathcal{T}^{-1}$ represents the inverse contourlet transformation; $f_{h}^{fct,s}$/$f^{cnn,s}$ denotes the CNN features and contourlet features (subbands), respectively. The details of \textit{Spatial–Spectral Fusion Module} are illustrated in Fig. \ref{fig:net} (\textit{Spatial–Spectral Fusion Module}). In this way, we enhance the CNNs' ability of geometric transformation by composing an explicit panoramic structure with a multidirectional analysis tool (contourlet transform). Since the geometric details of the RGB image are redundant for the depth map, if \textit{Spatial–Spectral Fusion Module} is directly applied to the decoder stage, it will inevitably interfere with the depth prediction. To avoid the disturbance and further leverage the geometric cue in the decoder stage, we design the \textit{Coefficients Generation Module}. Specially, we employ the geometric prior from the \textit{Spatial–Spectral Fusion Module}, and we can sparse the coefficient maps purposefully. In contrast to~\cite{ramamonjisoa2021single}, our generated coefficient maps are more accurate, and the sparsification process does not degrade our model’s performance. 
\begin{algorithm}[tb]
\caption{Mask Generation}
\label{alg:algorithm}
\textbf{Input}: High-frequency coefficient maps from forward contourlet transform: $HF_{fct}^{s}$\\
\textbf{Parameter}: None\\
\textbf{Output}: Mask $M_{s}$ for scale s
\begin{algorithmic}[1] 
\STATE Set a non-overlapping sliding window with size of 7$\times$1,
\FOR{ each position in $HF_{fct}^{s}$}
\STATE    keep the maximum value in the window, and set others to 0
\ENDFOR
\STATE Set a non-overlapping sliding window with size of 1$\times$7,
\FOR{ each position in $HF_{fct}^{s}$}
\STATE    keep the maximum value in the window, and set others to 0
\ENDFOR
\STATE $\theta_{i} \longleftarrow$ calculate threshold for $i$ th coefficient map
\FOR{$i$ th coefficient map}
\IF {the absolute value of $HF_{fct}^{s, i}$ is less than $\theta$}
\STATE the same position in $M_{s}$ is set to 0.
\ELSE
\STATE the same position in $M_{s}$ is set to 1.
\ENDIF
\ENDFOR
\STATE \textbf{return} $M_{s}$
\end{algorithmic}
\end{algorithm}
\subsubsection{Coefficient Generation Module}
\label{CGM}
It can be observed that real high-frequency coefficient maps have a few non-zero values that are located around depth edges. It means the non-zero coefficients indicate the geometric cue. 
Moreover, the high-frequency coefficients of the depth map are more sparse than those of the RGB images. To prevent the ``noise" from the RGB geometric cue, we design \textit{Coefficients Generation Module} to make the sparse coefficient maps more sparse.
Specifically, we use the coefficient maps generated by the forward contourlet decomposition in the encoder part to generate a target matrix for erasing the excess learned non-zero coefficients. For each scale, we generate a mask by detecting the local maximum value. Then to further generate more accurate masks, we sparse the generated coefficient masks with VisuShrink threshold function \cite{zhao2016scanner}. The threshold $\theta$ is calculated with Eq. \ref{thresh}.
\begin{equation}
\label{thresh}
\theta = \frac{MAD\sqrt{2\ln{N}}}{0.6745} 
\end{equation}
where $MAD$ is the median of the absolute coefficient matrix, $N$ is the number of coefficients. The whole mask generating method is described in \textit{Algorithm 1}. And the final learned high-frequency coefficient maps $HF_{s}$ are expressed as:
\begin{eqnarray}
HF_{s} = M_{s}\odot HF_{s}^{cnn}
\end{eqnarray}
where $M_{s}$ is the generated mask with scale s and $HF_{s}^{cnn}$ is the learned high-frequency coefficient maps, and $\odot$ represents element-wise multiply. Actually, we sparse the masks in two steps: generate a mask by detecting the maximum local value and further sparse the mask with the VisuShrink threshold function. Since significant wavelet coefficients tend to be concentrated at depth discontinuities, we preserve significant local coefficients by detecting the maximum local value. However, the significant local coefficients generated in this way are still redundant for the depth map, so we employ VisuShrink Function for further processing.
\begin{figure}[t]
\centering
\includegraphics[width=0.45\textwidth]{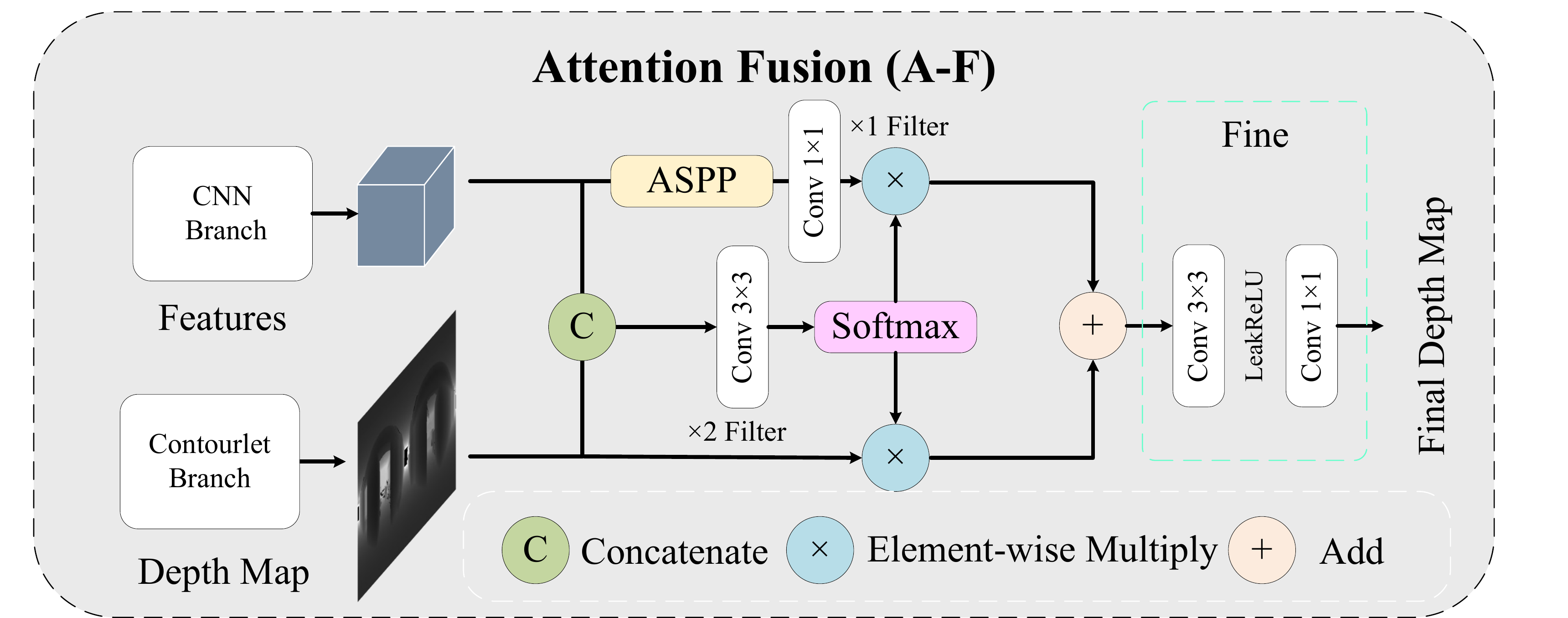}
\caption{Attention Fusion. We employ attention on fusion module to fuse the two branches.}
\label{fig:AF}
\end{figure}
\subsubsection{Attention Fusion Module}
\label{AF}
We propose an attention fusion module to fuse the dual-branch for a final depth map. As shown in Fig. \ref{fig:AF}, we concatenate the depth map with the CNN features to get two attention maps through convolution with the filter number set to 2 followed by the softmax function. Before fusing with the learned attention map in CNN branch, we employ a \textit{Atrous Spatial Pyramid Pooling}~\cite{chen2014semantic} block to establish a global perception field and a fully connected layer to get a depth map with dimension 1. Then we fuse the two depth maps with the learned attention wights. Finally, two convolutional layers are employed to refine the final depth map.  
\subsection{Objective Function}
\label{objective}
For better supervision, we combine reverse Huber (or Berhu) ~\cite{laina2016deeper,2020BiFuse,Pintore_2021_CVPR, Jiang_2021} loss and gradient loss~\cite{shen2021distortion} to design our objective function as commonly used in previous works~\cite{shen2021distortion,Pintore_2021_CVPR}. The standard inverse Huber loss is:
\begin{eqnarray}
\beta=\left
\{\begin{array}{ll}
|x|, & |x| \leq c \\
\frac{x^{2}+c^{2}}{2 c}, & |x|>c
\end{array} \right.
\end{eqnarray}
And in our objective function, the Berhu loss can be written as:
\begin{small}
\begin{eqnarray}
\beta_{\delta}(gt, pred)=\left\{\begin{array}{ll}
|gt-pred| & \text { for }|gt-pred| \leq \delta \\
\frac{|(gt-pred)^{2}|+ \delta^{2}}{2\delta} & \text { otherwise }
\end{array}\right.
\end{eqnarray}
\end{small}
where $gt$ and $pred$ represent ground truth and the predicted values, respectively; and in this paper, $\delta$ = 0.2.
\begin{table*}[t]
\begin{center}
\caption{\textsc{Quantitative comparison on the four popular datasets. *We retrain the model using the released code for fair comparison.$^{\dag}$}We re-evaluate the model using the released trained model for fair comparison. }
\label{table:c1}
\begin{tabular}{ccccccccc}
\hline\noalign{\smallskip}
Dataset&Method & MRE$\downarrow$ & MAE$\downarrow$ & RMSE$\downarrow$ & RMSElog$\downarrow$ & $\delta_{1}\uparrow$ & $\delta_{2}\uparrow$ & $\delta_{3}\uparrow$\\
\noalign{\smallskip}
\hline
\hline
&FCRN\cite{laina2016deeper}       & 0.2409  & 0.4008 & 0.6704 & 0.1244 & 0.7703 &0.9174 &0.9617     \\
&OmniDepth\cite{zioulis2018omnidepth} & 0.2901  & 0.4838 & 0.7643 & 0.1450 & 0.6830 & 0.8794 & 0.9429      \\
&Bifuse\cite{2020BiFuse}    & 0.2048  & 0.3470 & 0.6259 & 0.1134 &0.8452 & 0.9319 & 0.9632     \\
Matterport3D&HoHoNet\cite{sun2021hohonet}   & 0.1488  & 0.2862 & 0.5138 & 0.0871 & 0.8786 & 0.9519 & 0.9771    \\
&SliceNet\cite{Pintore_2021_CVPR}   & 0.1764  & 0.3296 & 0.6133 & 0.1045 & 0.8716 & 0.9483 & 0.9716    \\
&Unifuse\cite{Jiang_2021} & -  & 0.2814 & 0.4941 & 0.0701 & 0.8897 & 0.9623 & 0.9831\\
&Ours   & \textbf{0.0786}  & \textbf{0.1471} & \textbf{0.3799} & \textbf{0.0469} & \textbf{0.9048} & \textbf{0.9786} & \textbf{0.9926}    \\
\hline
\noalign{\smallskip}
\hline
&FCRN\cite{laina2016deeper}       & 0.1837  & 0.3428 & 0.5774 & 0.1100 & 0.7230 &0.9207 &0.9731     \\
&OmniDepth\cite{zioulis2018omnidepth} & 0.1996  & 0.3743 & 0.6152 & 0.1212 & 0.6877 & 0.8891 & 0.9578      \\
&Bifuse\cite{2020BiFuse}    & 0.1209  & 0.2343 & 0.4142 & 0.0787 &0.8660 & 0.9580 & 0.9860     \\
Stanford2D3D&HoHoNet\cite{sun2021hohonet}   & 0.1014  & 0.2027 & 0.3834 & 0.0668 & 0.9054 & 0.9693 & 0.9886    \\
&Unifuse\cite{Jiang_2021} & -  & 0.2082 & 0.3691 & 0.0721 & 0.8711 & 0.9664 & 0.9882\\
&SliceNet$^{\dag}$\cite{Pintore_2021_CVPR}   & 0.0744  & 0.1048 & 0.3684 & 0.0823 & 0.9029 & 0.9626 & 0.9844    \\
&Ours   & \textbf{0.0558}  & \textbf{0.0982} & \textbf{0.3528} & \textbf{0.0339} & \textbf{0.9140} & \textbf{0.9720} & \textbf{0.9903}    \\
\hline
\noalign{\smallskip}
\hline
&SliceNet*\cite{Pintore_2021_CVPR}       & 0.0743  & 0.0879 & 0.1231 & 0.0268 & 0.9529 &0.9826 &0.9908     \\
Structured3D&HoHoNet*\cite{sun2021hohonet} & 0.0457  & 0.0510 & 0.0872 & 0.0197 & 0.9713 & 0.9898 & 0.9947\\
&Unifuse*\cite{Jiang_2021} & 0.0465  & 0.0587 & 0.0892 & 0.0184 & 0.9788 & 0.9922 & 0.9956      \\
&Ours   & \textbf{0.0376}  & \textbf{0.0451} & \textbf{0.0801} & \textbf{0.0173} & \textbf{0.9809} & \textbf{0.9933} & \textbf{0.9962}    \\
\hline
\end{tabular}
\end{center}
\end{table*}

To obtain depth edges, we use eight convolution kernels to obtain gradients in eight directions (note that in this paper, we employ 8 ($2^{3}$)  direction contourlet transformation). They are represented as $G_{k}^{i}, i \in \{1, 2, 3, 4, 5, 6, 7, 8\}$ (where\begin{eqnarray}
G_{k}^{1} = [[-1., -2., -1.], [0., 0., 0.], [1., 2., 1.]]; \\G_{k}^{2} = [[-2., -1., 0.], [-1., 0., 1.], [0., 1., 2.]]; \\G_{k}^{3} = [[-1., 0., 1.], [-2., 0., 2.], [-1., 0., 1.]]; \\G_{k}^{4} = [[0., 1., 2.], [-1., 0., 1.], [-2., -1., 0.]]; \\G_{k}^{5} = [[1., 2., 1.], [0., 0., 0.], [-1., -2., -1.]]; \\G_{k}^{6} = [[2., 1., 0.], [1., 0., -1.], [0., -1., -2.]]; \\G_{k}^{7} = [[1., 0., -1.], [2., 0., -2.], [1., 0., -1.]]; \\G_{k}^{8} = [[0., -1., -2.], [1., 0., -1.], [2., 1., 0.]]
\end{eqnarray}). And the gradient of a panorama $E$ can be written as:
\begin{eqnarray}
8-Grad(E) = \sum_{i=1}^{8} G_{k}^{i}(E)
\end{eqnarray}
Due to the depth reconstructed level by level for a full-resolution, we need to supervise the estimated depth map for each scale. We apply the inverse contourlet transform to the ground truth, obtaining 4 scale depth maps to supervise the outputs from the inverse contourlet transform branch. The expression is:

\begin{eqnarray}
I_{s}, \Omega_{s} &=& \mathcal{T}^{-1}(I_{s-1}), s\in{1, 2, 3, 4}
\end{eqnarray}

where $I_{0}$ represent the input image, $I_{s}, s\in{1, 2, 3, 4}$ are the low-frequency parts, and $\Omega_{s}, s\in{1, 2, 3, 4}$ are the high-frequency parts. $\mathcal{T}^{-1}$ represents the inverse contourlet transform.

Hence, in our objective function, the total loss can be
\begin{equation}
\begin{split}
\ell_{total} &= 0.2(\beta_{0.2} (gt_{0},pr_{0})\\& + \beta_{0.2} (8-Grad(gt_{0}), 8-Grad(pr_{0})) \\&+ 0.8(\beta_{0.2} (gt_{0},pr_{c}) \\&+ \beta_{0.2} (8-Grad(gt_{0}), 8-Grad(pr_{c})) \\&+ 0.1\sum_{s=1}^{4}\beta_{0.2} (\mathcal{T}^{-1}(gt_{s-1}),pr_{s})
\end{split}
\end{equation}
where $gt_{0}, pr_{0}, pr_{c}$ denote the ground truth and final predicted values, and predict values from contourlet branch, respectively.
\begin{figure}[H]
\centering
\includegraphics[width=0.45\textwidth]{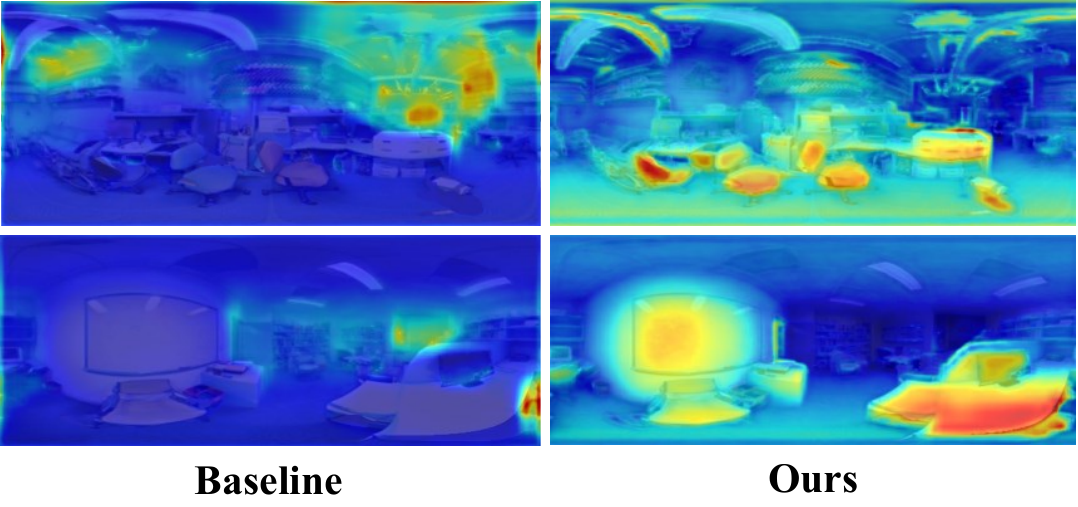}
\caption{Visualization of feature maps. We visualize the feature maps from the third layer of the decoder to compare the attention distributions.}
\label{fig:silence}
\end{figure}
\subsection{Interpretable analysis}
\label{Interpretable}
The lack of interpretability hinders the application of monocular depth estimation to downstream tasks (e.g., autonomous driving)~\cite{you2021towards}. Most existing solutions devote to achieving more advanced performance but ignore the model’s interpretability. Or they just backward inference how the model works according to its advanced performance. Even You et al.~\cite{you2021towards} enhance the interpretability by the depth selectivity of the model's hidden units. But the process of selecting hidden units is based on observation, which is also a backward inference.

In contrast, our method directly captures an explicit geometric cue that is beneficial for depth estimation. Unlike previous works that heavily rely on the learning ability of deep networks to infer depth (data-driven), we know exactly how our network works (rely on geometric cues to infer depth~\cite{dijk2019neural}). For further illustration, we visualize the feature maps in Fig. \ref{fig:silence}. The figure shows that the attention distribution in the baseline feature maps is scattered, while our model can focus on crucial structural information guided by geometric cues. The interpretability makes our model reliable, which improves the generalization of the model.
\begin{figure*}[t]
\centering
\includegraphics[width=\textwidth]{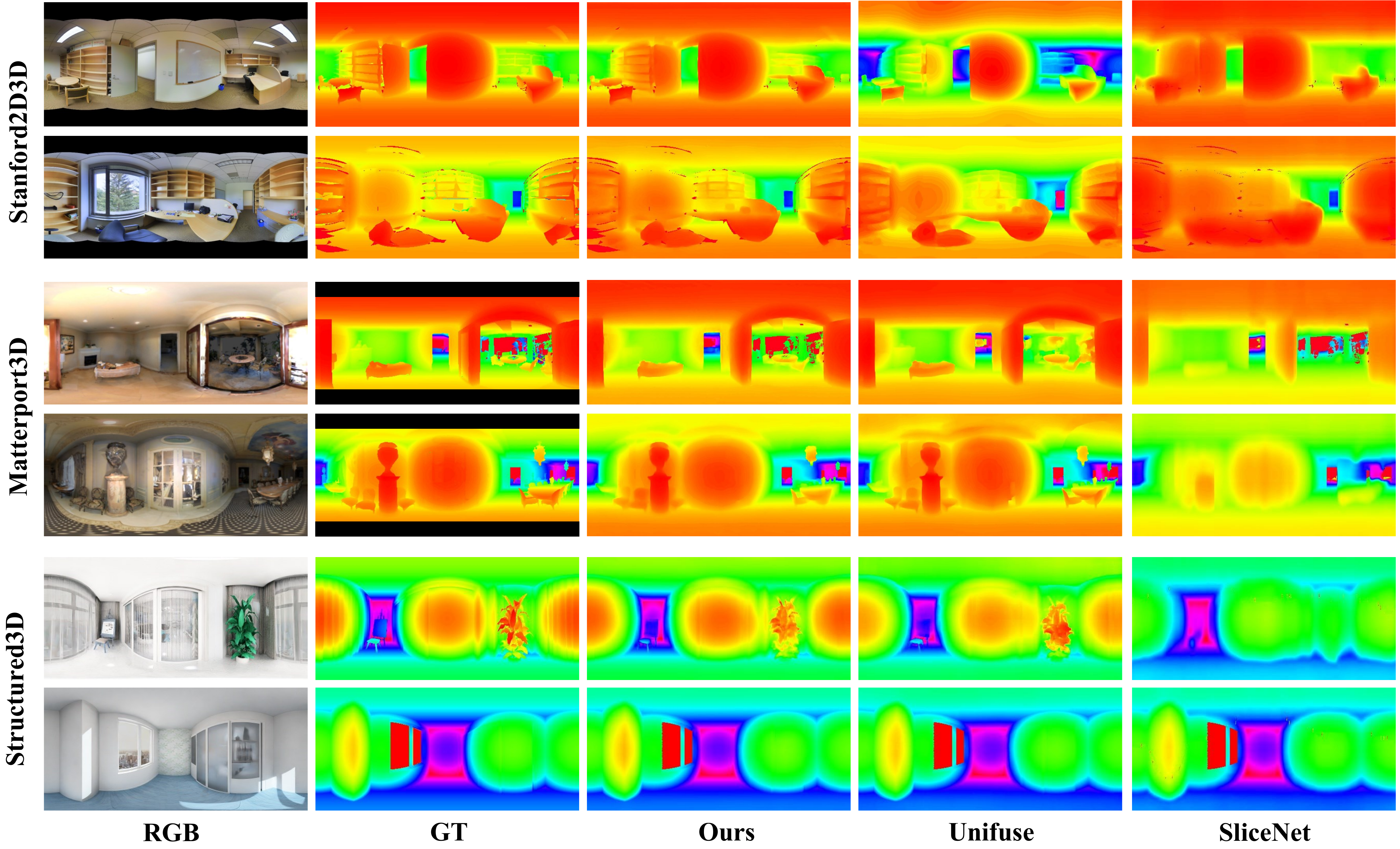}
\caption{Qualitative results. We compare our method with Unifuse\cite{Jiang_2021} and SliceNet\cite{Pintore_2021_CVPR}. A unified visualization method is adopted for all depths, and the closer the color and structure are to the ground truth, the better the prediction results.}
\label{fig:cc1}
\end{figure*}
\section{Experiment and Analysis}
\label{experiment}
In this section, we conduct extensive experiments on two publicly available real-world panoramic datasets (Matterport3D~\cite{Chang2018Matterport3D} and Stanford2D3D~\cite{DBLP:journals/corr/ArmeniSZS17}) and a high-quality synthesis dataset (Structured3D~\cite{zheng2020structured3d}) to verify the effectiveness of our proposed approach. Moreover, we perform a series of ablation studies to show the effectiveness of each module.  
 
 \subsection{Experimental Setup and Complexity}
 For a fair comparison, we strictly follow the previous works and use the official splits. The resolution of the input images is 512 × 1024 on the two real datasets and 256×512 on the synthesis dataset, respectively. For a fair comparison, we retrain three current state-of-the-art methods (Unifuse,  SliceNet and HoHoNet) on Structured3D using the released code. We follow the commonly used evaluation protocol, where the depth is clipped to 10 meters, there is no depth median alignment, and illegal values are masked out. We implement our method using Pytorch 1.7.1. We trained the proposed neural contourlet network on an NVIDIA RTX 3090 GPU with a batch size of 4 and a learning rate of 0.0001, using the Adam optimizer and keeping the default settings. We choose ResNet50 pretrained on ImageNet as the backbone of our network, the appropriate epoch is around 120 for Stanford2D3D and 40 for Matterport3D and 10 for Structured3D. Single-GPU inference time is 64ms (13 fps, same setup as SliceNet) for a 1024×512 image, with only 60M parameters. It is important to note that our network has fewer parameters and obtains a faster inference speed than the SOTA solutions (e.g. \textbf{64ms (ours)} vs. 74ms (SliceNet) in inference speed and \textbf{60M (ours)} vs. 75M (SliceNet) in parameters with a 1024×512 image).
 \subsection{Quantitative and Qualitative Evaluations}
 Similar to the previous works, we estimate our method with the commonly used error metrics: MAE, MRE, RMSE, RMSE log scale invariant, and $\delta_{t}$, $t\in {1.25, 1.25^2, 1.25^3}$. Specially, the metrics of dbe$^{acc}$, dbe$^{comp}$, prec$_{t}$, and rec$_{t}$, $t\in {0.25, 0.5, 1}$ provided by Pano3D~\cite{albanis2021pano3d} are employed to evaluate the capability of boundary preservation of models.
  
 \subsubsection{Quantitative analysis.} Tab. \ref{table:c1} illustrate our quantitative results on Stanford2D3D, Matterport3D, and Structured3D. In the tables, we compare with the most current SOTA approaches (e.g. Bifuse\cite{2020BiFuse}, HoHoNet\cite{sun2021hohonet}, Unifuse\cite{Jiang_2021}, SliceNet\cite{Pintore_2021_CVPR}).
 
Our method outperforms the other solutions in terms of all metrics, bolded in Tab. \ref{table:c1}. In particular, the RMSE metric of our model achieves a 23\% improvement on Matterport3D. However, the performance improvement of our model is relatively small on Stanford2D3D. As described in \cite{Jiang_2021}, we also find that some depth in Stanford2D3D is inaccurate, which degrades our model's ability to perceive an explicit structure, resulting in a slight improvement. (Note that the differences in calculating RMSE\footnote{\url{https://github.com/crs4/SliceNet}} lead to unfair comparisons, especially on the Stanford2D3D dataset. So we re-evaluate the metric of RMSE.)

 In addition, we can observe that the results of our model on the two real-world datasets are analogous. It shows that our model has better robustness. Because the two datasets have similar scenarios, the difference is that Stanford2D3D is smaller in scale. But clearly, our method is not affected by the scale of the dataset. Because our proposed \textit{Spatial–Spectral Fusion Module} provides a reliable panoramic structure with strong priors, it reduces the difficulty of network training and improves the accuracy of scene detail depth prediction.
 \subsubsection{Qualitative analysis.} Fig. ~\ref{fig:cc1} shows the qualitative comparison with the current SOTA approaches. From the figures, we can observe that our results are more competitive with clearly visible depth edges, which indicates that our network makes good use of panoramic structural cues. Furthermore, our attention to structural cues makes our network get rid of the negative effect of panoramic distortions: our method can more accurately predict depth, in large and small distorted regions (Fig. \ref{fig:cc1}). However, we can see that SliceNet's depth prediction for gravity-aligned regions is relatively accurate, but loses a lot of detail. It is worth noting that there are many daily necessities and furniture in indoor scenes, which require stronger network feature extraction capabilities to cover scene\begin{table}[t]
\begin{center}
\caption{\textsc{Generalization Comparison. Following Pano3D, we validate the generalization of our model by training on Matterport3D and testing on GibsonV2.}}
\label{table:gene}
\setlength\tabcolsep{3pt}
\begin{tabular}{c c c c c}
\hline
GV2(HR) & Method & RMSE $\downarrow$ & AbsRel $\downarrow$ & $\delta_{1}\uparrow$\\
 \hline
 & UNet$^{vnl}$\cite{albanis2021pano3d}  & 0.5794 & 0.2151& 0.6205\\
$tiny$ & ResNet$_{skip}^{comb}$\cite{albanis2021pano3d}  & 0.4993 & 0.1758 & 0.8031\\
& Ours  & \textbf{0.3910} & \textbf{0.1578} & \textbf{0.8145} \\
\hline
\noalign{\smallskip}
\hline
 & UNet$^{vnl}$\cite{albanis2021pano3d}  & 0.5901 & 0.2269 & 0.6102\\
$medium$& ResNet$_{skip}^{comb}$\cite{albanis2021pano3d}  & 0.4528 & 0.1664 & 0.8191\\
& Ours  & \textbf{0.3625} & \textbf{0.1543} & \textbf{0.8201} \\
\hline
\noalign{\smallskip}
\hline
& UNet$^{vnl}$\cite{albanis2021pano3d}  & 0.8772 & 0.2730 & 0.4609\\
$fullplus$ & ResNet$_{skip}^{comb}$\cite{albanis2021pano3d}  & 0.6607 & 0.1836 &0.7488\\
& Ours  & \textbf{0.5034} & \textbf{0.1697} & \textbf{0.7523} \\
\hline
\end{tabular}
\end{center}
\end{table}
details. Furthermore, SliceNet has poor anti-jamming ability because it works on the premise that the camera must be placed horizontally. It can be seen from the results that Unifuse has a large depth prediction error in large distortion regions. In contrast, reliable panoramic geometric cues can tolerate the distortion well, helping to find more accurate depth even in regions with repetitive structural details.

\subsubsection{Generalizability.}
\begin{table*}[t]
\begin{center}
\caption{\textsc{Boundary preserve comparison.}}
\label{table:bp}
\begin{tabular}{cccccccccc}
\hline\noalign{\smallskip}
Dataset&Method & dbe$^{acc}\downarrow$ & dbe$^{comp}\downarrow$ & prec$_{0.25}\uparrow$ & prec$_{0.5}\uparrow$ & prec$_{1}\uparrow$ & rec$_{0.25}\uparrow$ & rec$_{0.5}\uparrow$ & rec$_{1}\uparrow$\\
\hline\noalign{\smallskip}
\hline
&Pnas$^{comb}$\cite{albanis2021pano3d}       & 2.5119  & 5.3501 & 0.3983 & 0.3159 & 0.2701 &0.2353 &0.1442 &0.1098     \\
&UNet$^{vnl}$\cite{albanis2021pano3d} & 1.2699  & 3.8876 & 0.5897 & 0.5754 & 0.5185 &0.4396 &0.3669 &0.2859      \\
Matterport&DenseNet$^{comb}$\cite{albanis2021pano3d}    & 2.0628  & 5.0977 & 0.4716 & 0.4077 & 0.3520 &0.2609 &0.1687 &0.1221     \\
3D&ResNet$^{comb}$\cite{albanis2021pano3d}    & 2.2393  & 5.3796 & 0.4410 & 0.3670 & 0.2744 &0.2291 &0.1223 &0.0720    \\
&ResNet$^{comb}_{skip}$\cite{albanis2021pano3d} & 1.4883  & 4.5346 & 0.5734 & 0.5411 & 0.4757 &0.3399 &0.2430 &0.1637\\
&Spherical-ResNet & 1.6924  & 2.8681 & 0.6275 & 0.4935 & 0.3603 &0.5475 &0.4581 &0.3796\\
&Ours   & \textbf{1.0146} & \textbf{2.6699}  & \textbf{0.6886} & \textbf{0.5913} & \textbf{0.5200} & \textbf{0.5886} & \textbf{0.5012} & \textbf{0.4144}    \\
\hline
\end{tabular}
\end{center}
\end{table*} 
We validate the generalization capability by training our model on Matterport3D and testing on GibsonV2 ~\cite{Xia_2018_CVPR} (splits of tiny, medium, and fullplus, strictly follow the protocol of Pano3D \cite{albanis2021pano3d}). The GibsonV2 dataset is more extensive than Matterport3D and offers a wider variety of scenarios. The results are exhibited in Tab. \ref{table:gene}. Our solution achieved the best performance when transferred to GibsonV2. Other existing methods are data-driven and heavily rely on the network’s learning capability. In contrast, our contourlet branch can extract geometric cues explicitly, which helps the monocular depth estimation task. In addition, the contourlet transform can provide the local  features extraction capability with sufficient receptive field on the geometric structures,
enhancing the network’s learning capability. Hence, our method has better generalizability.
 
 \subsubsection{Convergence.}
\begin{figure}[t]
\centering
\includegraphics[width=0.45\textwidth]{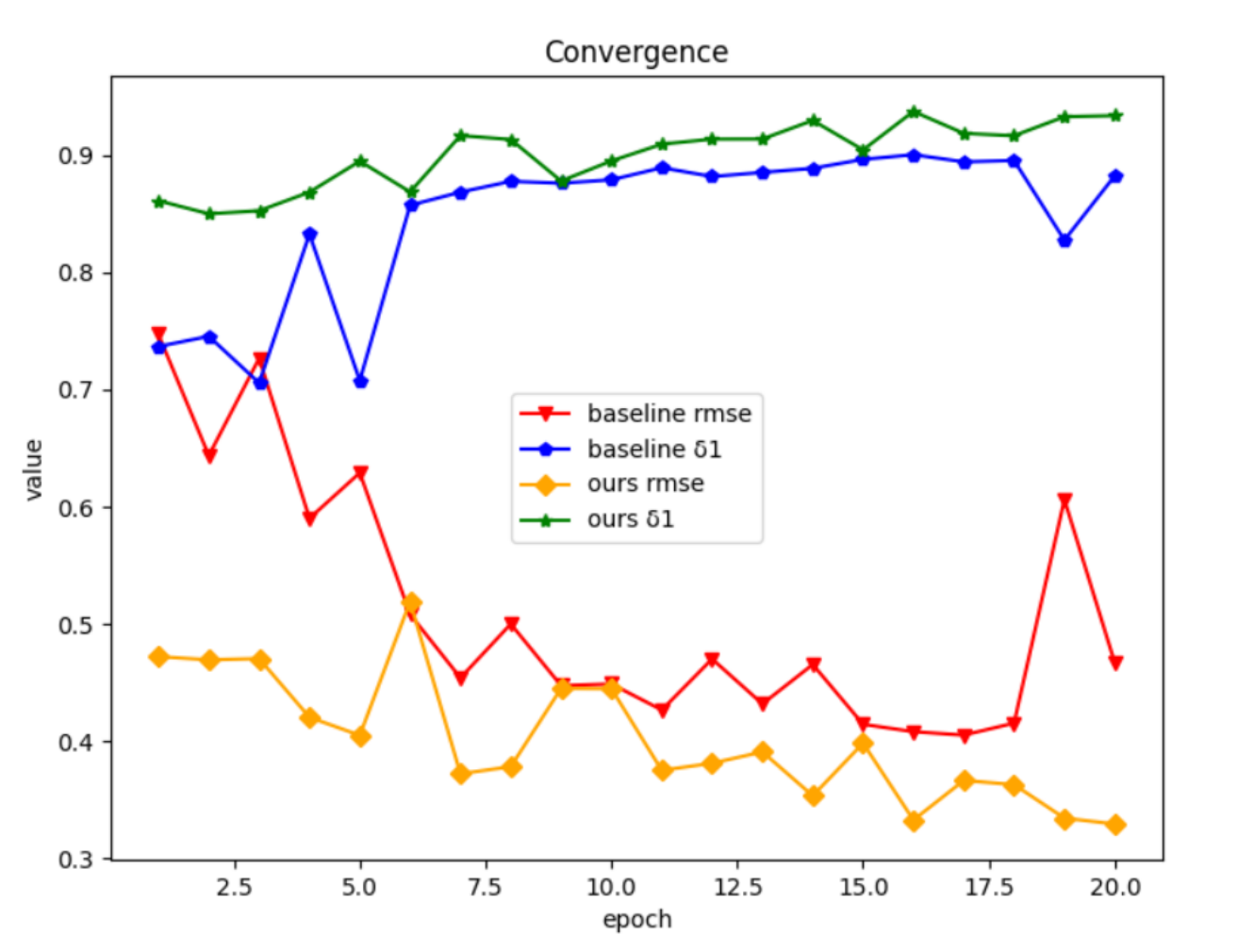}
\caption{Convergence comparison with the baseline.}
\label{fig:conver}
\end{figure}

 To compare the convergence, we visualize the changes in RMSE and $\delta_{1}$ over the first twenty epochs and exhibit them in Fig. \ref{fig:conver}. We can observe that our model achieves better performance after one epoch of training, which reflects the fast and stable convergence capacity of our model.
\begin{table*}[t]
\begin{center}
\caption{\textsc{Ablation study. We conduct a series of ablation studies on Stangford2D3D, a challenge dataset (small scale, real scenes).}}
\label{table:abla}
\begin{tabular}{c c c c c c c c c c c c c c}
\hline
&Baseline & SSFM & CGM & Contourlet & Haar &A-F& Peak-Mask & 8-Grad &Params & Speed & RMSE & $\delta_{1}$\\
\hline
a&\checkmark &   &  &  & &&& & 54M & 21ms & 0.4253 & 0.8960     \\
b&\checkmark & \checkmark  &  & \checkmark && && & 56M & 43ms & 0.3670 & 0.9214      \\
c&\checkmark & \checkmark  & \checkmark & \checkmark&&& & &60M & 64ms & 0.3266 & 0.9252\\
d&\checkmark & \checkmark & \checkmark &  & \checkmark &&& & 56M & 25ms & 0.3729 & 0.9041    \\
e&\checkmark & \checkmark  & \checkmark & \checkmark& &\checkmark && &60M & 64ms & 0.3052 & 0.9453\\
f&\checkmark & \checkmark  & \checkmark & \checkmark& &\checkmark &\checkmark& &60M & 64ms & 0.2952 & 0.9525\\
g&\checkmark & \checkmark  & \checkmark & \checkmark& &\checkmark &\checkmark & \checkmark&60M & 64ms & \textbf{0.2789} & \textbf{0.9614}\\
h&\multicolumn{8}{c}{Cube projection with haar wavelet}&54M & 34ms & 0.4041 & 0.9058\\
\hline
\end{tabular}
\end{center}
\end{table*}
\begin{figure*}[t]
\centering
\includegraphics[width=0.9\textwidth]{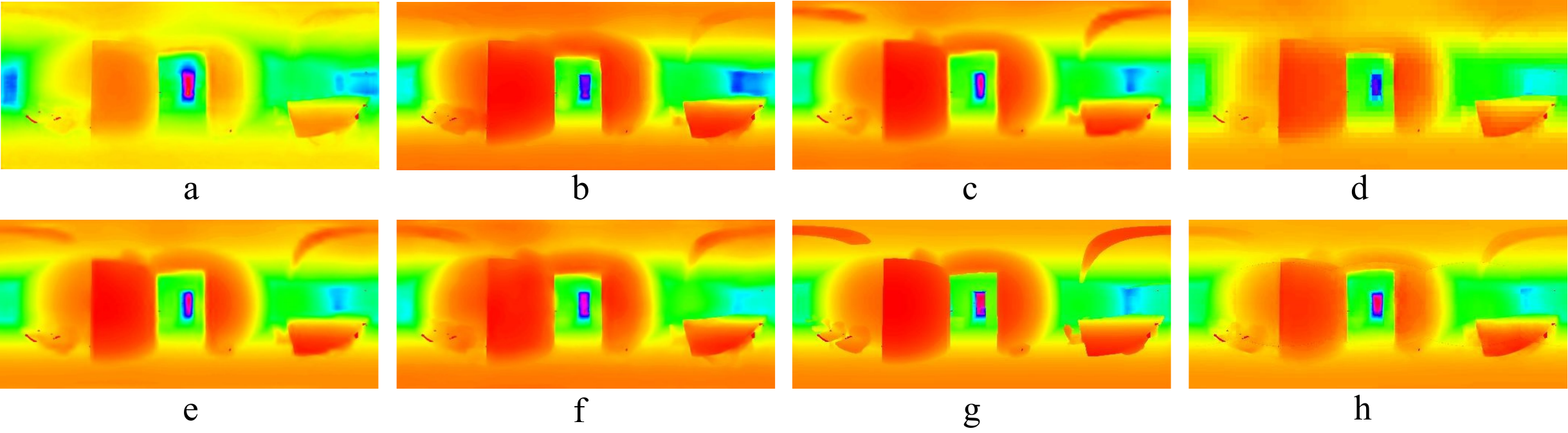}
\caption{Example of qualitative results of ablation study. a) is the baseline; b) means adding the \textit{Spatial–Spectral Fusion Module}  based on baseline, and \textit{Spatial–Spectral Fusion Module} is implemented by using contourlet; c) means continue to add \textit{Coefficients Generation Module}; d) represents that the part of the wavelet transform in the network is completely replaced by the Haar Wavelet. e) is further added with attention fusion module; f) uses the proposed mask generation method; g) employs the proposed 8 directional gradient; h) employs the strategy that described in Sec.~\ref{CvsH}. A unified visualization method is adopted for all depths, and the closer the color and structure are to the ground truth, the better the prediction results.}
\label{fig:abla}
\end{figure*}
\subsubsection{Boundary preservation.}
To quantitatively evaluate our model's boundary preservation capability, we strictly follow the protocol of Pano3D~\cite{albanis2021pano3d} and employ the metrics they provided. The results are reported in Tab. \ref{table:bp}. From the table, we can observe that the boundary preservation ability of our model is much better than the baseline provided in Pano3D~\cite{albanis2021pano3d}. Actually, the baseline only employs a gradient loss in the objective function for supervision without any other structure-specific designs, resulting in a limited boundary preservation capability. In contrast, we not only use contourlet to perceive explicit panoramic structure but also design a panoramic-suitable gradient loss in the objective function. The two strategies can all contribute to preserving a better boundary in the predicted depth maps. 
Besides, we design Spherical-ResNet by replacing the standard convolution layers with spherical convolution layers~\cite{cohen2018spherical}, and present the results in Tab. \ref{table:bp}. Spherical-ResNet improves the performance of the boundary preservation metrics compared to the baseline and achieves an RMSE of 0.3983. However, it just reaches 0.8431 on $\delta_{1}$. In fact, The boundary preservation capability of Spherical-ResNet is learned during the training stage. It is data-driven and implicit, and the learned structure may be distracted by other learned features. Whereas our method directly extracts an explicit geometric structure and leverages it as a cue to enhance depth estimation. Hence, our approach is more solid and has better performance. Besides, spherical convolution is computationally intensive, causing more computational memory (e.g., an NVIDIA RTX 3090 GPU can only afford one sample with a 1024*512 resolution in a batch).

 \subsection{Ablation Study}
 \label{abla}
 We conduct an ablation study to verify the effectiveness of each module. Baseline is an encoder-decoder structure composed of ResNet50 as an encoder and DenseNet decoder. ResNet-50 has been shown to have strong feature extraction capabilities and is widely used as a backbone for various vision tasks. The decoder part of DenseNet uses a combination of interpolation up-sampling and convolutional layers. For the Matterport3D and Stanford2D3D datasets, this combination favors depth estimation for gravity-aligned regions. To effectively evaluate the contribution of each part, we conduct experiments under the same settings, training on the Stanford2D3D dataset (provided by Zioulis~\cite{zioulis2018omnidepth}, for its integrity, high quality, dense annotation and the smaller scale) without any pre-trained weights.

\subsubsection{Effectiveness of \textit{Spatial–Spectral Fusion Module}.} The purpose of \textit{Spatial–Spectral Fusion Module} is to enhance encoder feature extraction for panoramas by fusing reliable panorama structure from contourlet transform and semantic features from CNN branches. From Tab. \ref{table:abla}, we can observe that the \textit{Spatial–Spectral Fusion Module} improves RMSE by 12\% and $\delta_{1}$ by 3\%. Obviously, the RMSE improvement is more significant, which is mainly reflected in the sharper depth edge. The RMSE improvement is more significant, mainly reflected in sharper depth edges. From Fig. \ref{fig:abla}, we can intuitively see that the \textit{Spatial–Spectral Fusion Module}  makes the structure in the depth map clearer, and the predicted depth is closer to the ground truth. It also verifies that structural cues are important for neural networks to predict monocular depth maps.
\subsubsection{Effectiveness of \textit{Coefficients Generation Module}.} We further propose \textit{Coefficients Generation Module} to highlight structural cues in the network decoder part. Most previous similar works have overlooked this. The decoder part directly outputs the predicted depth value, and the reliable structural information can make the depth edge sharper. It is confirmed in Fig. \ref{fig:abla} (especially in group 1), where the \textit{Coefficients Generation Module} makes the edges in the prediction map clearer, even in large distorted regions. From Tab. \ref{table:abla}, we also can note that RMSE is improved by 11\%. Therefore, it is meaningful to enhance the structural cues of the monocular depth estimation network at all stages, mainly because it provides sharper depth edges.
\subsubsection{Effectiveness of A-F and 8-Grad Loss.}
\begin{figure}[!h]
\centering
\includegraphics[width=0.45\textwidth]{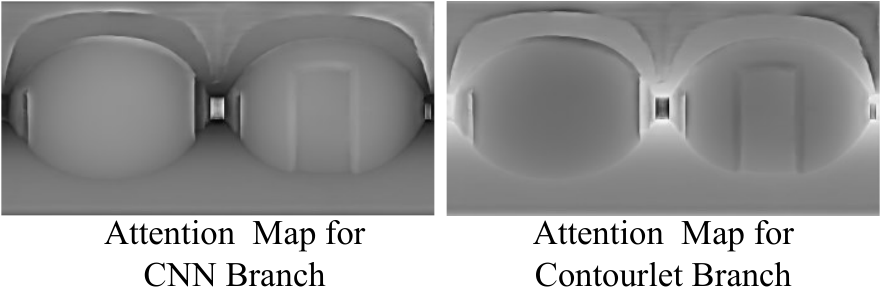}
\caption{Visualize the two attention maps from softmax function. We can see that the map for contourlet branch  focuses more on depth edges.}
\label{fig:AFdistribute}
\end{figure}

From Fig. \ref{fig:AFdistribute}, the learned attention maps have promising complementarity, further confirming the importance of structural and semantic cues. We can notice that the contourlet branch focuses more on structure prediction, while the semantic branch focuses more on content prediction. The 8-Grad loss corresponds to the eight directional subbands of the contourlet, which can make it perform better. 
\subsubsection{Contourlet vs. Haar.}\label{CvsH} We also verify the superiority of the contourlet transform for panoramas by replacing the contourlet transform with the haar wavelet. From Tab. \ref{table:abla}, we can observe that our framework with the Haar wavelet also outperforms the baseline 12\% on RMSE and 1\% on $\delta_{1}$. However, compared with contourlet transform, haar wavelet has obvious disadvantages. In Tab. \ref{table:abla}, contourlet transform outperforms Haar Wavelet 12\% on RMSE. This difference is mainly due to the poor ability of wavelet transform to approximate the curves. Because in the wavelet transform, the direction of the high-frequency components of its hidden structural features is limited (only horizontal, vertical, and diagonal directions), while the diversity of directions in the contourlet transform makes it more sensitive to curves. The mosaic effect (mainly at the edge of depth) in the depth map in Fig. \ref{fig:abla} is caused by the square support space in the wavelet.

We also attempt another strategy. To deal with the distortion in panoramas, we project the input ERP images to cube images and then apply a Haar wavelet to catch the geometric cue. Finally, we project the cube maps back to ERP format. The quantitative and qualitative results are exhibited in Tab. \ref{table:abla} and Fig. \ref{fig:abla}h, respectively. From the figure, we can observe that there are still visible stitching seams from cube maps (even though we employ spherical padding~\cite{2020BiFuse}). 

\section{Conclusions}
In this paper, we propose a simple but effective neural network suitable for monocular panoramic depth estimation, named the neural contourlet network. In detail, we design a bi-branch to capture an explicit geometric cue and an implicit cue, respectively. In view of the ``panorama-customized" properties of contourlet transform, we introduce it to extract a panorama geometric cue. To bridge the gap between spatial and spectral domain, we design a spatial-spectral fusion module to integrate the two cues in the encoder stage. We further propose a coefficient-generation module in the decoder stage to catch the geometric cue. Precisely, we design a mask generation algorithm to guide the generation of contourlet coefficients for omnidirectional depth. More specifically, we capture geometric cues in the decoder stage by adopting the coefficients maps generated from forward contourlet decomposition in the encoder part to get the masks. Ablation experiments verify the effectiveness of our proposed modules, and it proves that the contourlet is suitable for panoramic scenes. Experiments on three panoramic datasets show that our method outperforms other state-of-the-art methods. We hope our solution can inspire efforts in the field to explore effective and compact neural networks for monocular panoramic depth estimation. Furthermore, it should be noted that the power of traditional methods cannot be ignored and is beneficial to the neural network's learning ability, convergence speed, and parameter size.
\normalem
\bibliographystyle{ieeetr}
\bibliography{reference}

\begin{thebibliography}{10}

\bibitem{8353313}
H.~Kumar, A.~S. Yadav, S.~Gupta, and K.~S. Venkatesh, ``Depth map estimation
  using defocus and motion cues,'' {\em IEEE Transactions on Circuits and
  Systems for Video Technology}, vol.~29, no.~5, pp.~1365--1379, 2019.

\bibitem{9615228}
X.~Meng, C.~Fan, Y.~Ming, and H.~Yu, ``Cornet: Context-based ordinal regression
  network for monocular depth estimation,'' {\em IEEE Transactions on Circuits
  and Systems for Video Technology}, pp.~1--1, 2021.

\bibitem{8764412}
Y.~Cao, T.~Zhao, K.~Xian, C.~Shen, Z.~Cao, and S.~Xu, ``Monocular depth
  estimation with augmented ordinal depth relationships,'' {\em IEEE
  Transactions on Circuits and Systems for Video Technology}, vol.~30, no.~8,
  pp.~2674--2682, 2020.

\bibitem{8105853}
H.~Yan, X.~Yu, Y.~Zhang, S.~Zhang, X.~Zhao, and L.~Zhang, ``Single image depth
  estimation with normal guided scale invariant deep convolutional fields,''
  {\em IEEE Transactions on Circuits and Systems for Video Technology},
  vol.~29, no.~1, pp.~80--92, 2019.

\bibitem{dijk2019neural}
T.~v. Dijk and G.~d. Croon, ``How do neural networks see depth in single
  images?,'' in {\em Proceedings of the IEEE/CVF International Conference on
  Computer Vision}, pp.~2183--2191, 2019.

\bibitem{Chen_2021_CVPR}
X.~Chen, Y.~Wang, X.~Chen, and W.~Zeng, ``S2r-depthnet: Learning a
  generalizable depth-specific structural representation,'' in {\em Proceedings
  of the IEEE/CVF Conference on Computer Vision and Pattern Recognition
  (CVPR)}, pp.~3034--3043, June 2021.

\bibitem{shen2021distortion}
Z.~Shen, C.~Lin, L.~Nie, K.~Liao, and Y.~Zhao, ``Distortion-tolerant monocular
  depth estimation on omnidirectional images using dual-cubemap,'' in {\em 2021
  IEEE International Conference on Multimedia and Expo (ICME)}, pp.~1--6, IEEE,
  2021.

\bibitem{Pintore_2021_CVPR}
G.~Pintore, M.~Agus, E.~Almansa, J.~Schneider, and E.~Gobbetti, ``Slicenet:
  Deep dense depth estimation from a single indoor panorama using a slice-based
  representation,'' in {\em Proceedings of the IEEE/CVF Conference on Computer
  Vision and Pattern Recognition (CVPR)}, pp.~11536--11545, June 2021.

\bibitem{sivak1990integration}
B.~Sivak and C.~L. MacKenzie, ``Integration of visual information and motor
  output in reaching and grasping: the contributions of peripheral and central
  vision,'' {\em Neuropsychologia}, vol.~28, no.~10, pp.~1095--1116, 1990.

\bibitem{8307430}
J.~Yang, Y.~Xiao, and Z.~Cao, ``Aligning 2.5d scene fragments with distinctive
  local geometric features and voting-based correspondences,'' {\em IEEE
  Transactions on Circuits and Systems for Video Technology}, vol.~29, no.~3,
  pp.~714--729, 2019.

\bibitem{MengkunLiu2021CCNNCC}
M.~Liu, L.~Jiao, X.~Liu, L.~Lingling, F.~Liu, and S.~Yang, ``C-cnn: Contourlet
  convolutional neural networks,'' {\em IEEE Transactions on Neural Networks},
  vol.~32, pp.~2636--2649, 2021.

\bibitem{ramamonjisoa2021single}
M.~Ramamonjisoa, M.~Firman, J.~Watson, V.~Lepetit, and D.~Turmukhambetov,
  ``Single image depth prediction with wavelet decomposition,'' in {\em
  Proceedings of the IEEE/CVF Conference on Computer Vision and Pattern
  Recognition}, pp.~11089--11098, 2021.

\bibitem{DBLP:journals/corr/ArmeniSZS17}
I.~Armeni, S.~Sax, A.~R. Zamir, and S.~Savarese, ``Joint 2d-3d-semantic data
  for indoor scene understanding,'' {\em CoRR}, vol.~abs/1702.01105, 2017.

\bibitem{Chang2018Matterport3D}
A.~Chang, A.~Dai, T.~Funkhouser, M.~Halber, M.~Niebner, M.~Savva, S.~Song,
  A.~Zeng, and Y.~Zhang, ``Matterport3d: Learning from rgb-d data in indoor
  environments,'' in {\em 2017 International Conference on 3D Vision (3DV)},
  2018.

\bibitem{zheng2020structured3d}
J.~Zheng, J.~Zhang, J.~Li, R.~Tang, S.~Gao, and Z.~Zhou, ``Structured3d: A
  large photo-realistic dataset for structured 3d modeling,'' in {\em European
  Conference on Computer Vision}, pp.~519--535, Springer, 2020.

\bibitem{2020BiFuse}
F.~E. Wang, Y.~H. Yeh, M.~Sun, W.~C. Chiu, and Y.~H. Tsai, ``Bifuse: Monocular
  360 depth estimation via bi-projection fusion,'' in {\em 2020 IEEE/CVF
  Conference on Computer Vision and Pattern Recognition (CVPR)}, 2020.

\bibitem{Jiang_2021}
H.~Jiang, Z.~Sheng, S.~Zhu, Z.~Dong, and R.~Huang, ``Unifuse: Unidirectional
  fusion for 360° panorama depth estimation,'' {\em IEEE Robotics and
  Automation Letters}, vol.~6, p.~1519–1526, Apr 2021.

\bibitem{Eder2020Tangent}
M.~Eder, M.~Shvets, J.~Lim, and J.-M. Frahm, ``Tangent images for mitigating
  spherical distortion,'' in {\em 2020 IEEE/CVF Conference on Computer Vision
  and Pattern Recognition (CVPR)}, pp.~12423--12431, 2020.

\bibitem{Chen_2021}
H.-X. Chen, K.~Li, Z.~Fu, M.~Liu, Z.~Chen, and Y.~Guo, ``Distortion-aware
  monocular depth estimation for omnidirectional images,'' {\em IEEE Signal
  Processing Letters}, vol.~28, p.~334–338, 2021.

\bibitem{tateno2018distortion}
K.~Tateno, N.~Navab, and F.~Tombari, ``Distortion-aware convolutional filters
  for dense prediction in panoramic images,'' in {\em Proceedings of the
  European Conference on Computer Vision (ECCV)}, pp.~707--722, 2018.

\bibitem{eder2019mapped}
M.~Eder, T.~Price, T.~Vu, A.~Bapat, and J.-M. Frahm, ``Mapped convolutions,''
  {\em arXiv preprint arXiv:1906.11096}, 2019.

\bibitem{cohen2018spherical}
T.~S. Cohen, M.~Geiger, J.~K{\"o}hler, and M.~Welling, ``Spherical cnns,'' in
  {\em International Conference on Learning Representations}, 2018.

\bibitem{FernandezLabrador2020CornersFL}
C.~Fernandez-Labrador, J.~M. F{\'a}cil, A.~P{\'e}rez-Yus, C.~Demonceaux,
  J.~Civera, and J.~J. Guerrero, ``Corners for layout: End-to-end layout
  recovery from 360 images,'' {\em IEEE Robotics and Automation Letters},
  vol.~5, pp.~1255--1262, 2020.

\bibitem{Jin_2020_CVPR}
L.~Jin, Y.~Xu, J.~Zheng, J.~Zhang, R.~Tang, S.~Xu, J.~Yu, and S.~Gao,
  ``Geometric structure based and regularized depth estimation from 360 indoor
  imagery,'' in {\em Proceedings of the IEEE/CVF Conference on Computer Vision
  and Pattern Recognition (CVPR)}, June 2020.

\bibitem{zeng2020joint}
W.~Zeng, S.~Karaoglu, and T.~Gevers, ``Joint 3d layout and depth prediction
  from a single indoor panorama image,'' in {\em European Conference on
  Computer Vision}, pp.~666--682, Springer, 2020.

\bibitem{zioulis2022monocular}
N.~Zioulis, F.~Alvarez, D.~Zarpalas, and P.~Daras, ``Monocular spherical depth
  estimation with explicitly connected weak layout cues,'' {\em ISPRS Journal
  of Photogrammetry and Remote Sensing}, vol.~183, pp.~269--285, 2022.

\bibitem{8116695}
Y.~Chen, D.~Li, and J.~Q. Zhang, ``Complementary color wavelet: A novel tool
  for the color image/video analysis and processing,'' {\em IEEE Transactions
  on Circuits and Systems for Video Technology}, vol.~29, no.~1, pp.~12--27,
  2019.

\bibitem{QiufuLi2020WaveletIC}
Q.~Li, L.~Shen, S.~Guo, and Z.~Lai, ``Wavelet integrated cnns for noise-robust
  image classification,'' {\em arXiv: Computer Vision and Pattern Recognition},
  2020.

\bibitem{guo2017deep}
T.~Guo, H.~Seyed~Mousavi, T.~Huu~Vu, and V.~Monga, ``Deep wavelet prediction
  for image super-resolution,'' in {\em Proceedings of the IEEE conference on
  computer vision and pattern recognition workshops}, pp.~104--113, 2017.

\bibitem{deng2019wavelet}
X.~Deng, R.~Yang, M.~Xu, and P.~L. Dragotti, ``Wavelet domain style transfer
  for an effective perception-distortion tradeoff in single image
  super-resolution,'' in {\em Proceedings of the IEEE/CVF International
  Conference on Computer Vision}, pp.~3076--3085, 2019.

\bibitem{DavidLDonoho1998DataCA}
D.~L. Donoho, M.~Vetterli, R.~A. DeVore, and I.~Daubechies, ``Data compression
  and harmonic analysis,'' {\em IEEE Transactions on Information Theory},
  vol.~44, pp.~2435--2476, 1998.

\bibitem{he2016identity}
K.~He, X.~Zhang, S.~Ren, and J.~Sun, ``Identity mappings in deep residual
  networks,'' in {\em European conference on computer vision}, pp.~630--645,
  Springer, 2016.

\bibitem{he2019rethinking}
K.~He, R.~Girshick, and P.~Doll{\'a}r, ``Rethinking imagenet pre-training,'' in
  {\em Proceedings of the IEEE/CVF International Conference on Computer
  Vision}, pp.~4918--4927, 2019.

\bibitem{wang2018non}
X.~Wang, R.~Girshick, A.~Gupta, and K.~He, ``Non-local neural networks,'' in
  {\em Proceedings of the IEEE conference on computer vision and pattern
  recognition}, pp.~7794--7803, 2018.

\bibitem{huang2018condensenet}
G.~Huang, S.~Liu, L.~Van~der Maaten, and K.~Q. Weinberger, ``Condensenet: An
  efficient densenet using learned group convolutions,'' in {\em Proceedings of
  the IEEE conference on computer vision and pattern recognition},
  pp.~2752--2761, 2018.

\bibitem{ronneberger2015u}
O.~Ronneberger, P.~Fischer, and T.~Brox, ``U-net: Convolutional networks for
  biomedical image segmentation,'' in {\em International Conference on Medical
  image computing and computer-assisted intervention}, pp.~234--241, Springer,
  2015.

\bibitem{zhao2016scanner}
J.~Zhao, J.-S. Lee, H.~Xu, K.~Xu, Z.-H. Ren, J.-C. Chen, and C.-H. Wu,
  ``Scanner-dependent threshold estimation of wavelet denoising for
  small-animal pet,'' {\em IEEE Transactions on Nuclear Science}, vol.~64,
  no.~1, pp.~705--712, 2016.

\bibitem{chen2014semantic}
L.-C. Chen, G.~Papandreou, I.~Kokkinos, K.~Murphy, and A.~L. Yuille, ``Semantic
  image segmentation with deep convolutional nets and fully connected crfs,''
  {\em arXiv preprint arXiv:1412.7062}, 2014.

\bibitem{laina2016deeper}
I.~Laina, C.~Rupprecht, V.~Belagiannis, F.~Tombari, and N.~Navab, ``Deeper
  depth prediction with fully convolutional residual networks,'' in {\em 2016
  Fourth international conference on 3D vision (3DV)}, pp.~239--248, IEEE,
  2016.

\bibitem{zioulis2018omnidepth}
N.~Zioulis, A.~Karakottas, D.~Zarpalas, and P.~Daras, ``Omnidepth: Dense depth
  estimation for indoors spherical panoramas,'' in {\em Proceedings of the
  European Conference on Computer Vision (ECCV)}, pp.~448--465, 2018.

\bibitem{sun2021hohonet}
C.~Sun, M.~Sun, and H.-T. Chen, ``Hohonet: 360 indoor holistic understanding
  with latent horizontal features,'' in {\em Proceedings of the IEEE/CVF
  Conference on Computer Vision and Pattern Recognition}, pp.~2573--2582, 2021.

\bibitem{you2021towards}
Z.~You, Y.-H. Tsai, W.-C. Chiu, and G.~Li, ``Towards interpretable deep
  networks for monocular depth estimation,'' in {\em Proceedings of the
  IEEE/CVF International Conference on Computer Vision}, pp.~12879--12888,
  2021.

\bibitem{albanis2021pano3d}
G.~Albanis, N.~Zioulis, P.~Drakoulis, V.~Gkitsas, V.~Sterzentsenko, F.~Alvarez,
  D.~Zarpalas, and P.~Daras, ``Pano3d: A holistic benchmark and a solid
  baseline for 360° depth estimation,'' in {\em 2021 IEEE/CVF Conference on
  Computer Vision and Pattern Recognition Workshops (CVPRW)}, pp.~3722--3732,
  IEEE, 2021.

\bibitem{Xia_2018_CVPR}
F.~Xia, A.~R. Zamir, Z.~He, A.~Sax, J.~Malik, and S.~Savarese, ``Gibson env:
  Real-world perception for embodied agents,'' in {\em Proceedings of the IEEE
  Conference on Computer Vision and Pattern Recognition (CVPR)}, June 2018.

\end{thebibliography}
\end{document}